%% file: main.tex
\documentclass{article}

\PassOptionsToPackage{numbers, compress}{natbib}

\usepackage[preprint]{neurips_2023}

\usepackage{enumitem}
\usepackage[utf8]{inputenc} 
\usepackage[T1]{fontenc}    
\usepackage{hyperref}       
\usepackage{url}            
\usepackage{booktabs}       
\usepackage{amsfonts}       
\usepackage{nicefrac}       
\usepackage{microtype}      
\usepackage{xcolor}         

\usepackage{comment}
\usepackage{multirow}
\usepackage{makecell}
\usepackage[pdftex]{graphicx}
\usepackage{amsmath,amsfonts,amsthm}
\usepackage{mathtools}
\usepackage{adjustbox}
\usepackage{multicol}
\usepackage{multirow}
\usepackage[linesnumbered,ruled]{algorithm2e}
\usepackage{amsmath}
\usepackage{caption}
\usepackage{subcaption}
\usepackage{listings}
\usepackage{wrapfig}
\usepackage{tablefootnote}
\usepackage{tikz}
\usepackage{pifont}
\newcommand{\cmark}{\ding{51}}%
\newcommand{\xmark}{\ding{55}}%

\usepackage{changepage}

\theoremstyle{plain}

\theoremstyle{definition}

\theoremstyle{remark}

\title{Can Text-to-image Model Assist Multi-modal Learning for Visual Recognition with Visual Modality Missing?}

\author{%
  Tiantian Feng$^{1}$, Daniel Yang$^{1}$,Digbalay Bose$^{1}$, Shrikanth S. Narayanan$^{1}$ \\ 
\quad $^1$University of Southern California \\
{\small\texttt{tiantiaf@usc.edu}}
{
}
}

\begin{document}

\maketitle

\begin{abstract}
Multi-modal learning has emerged as an increasingly promising avenue in vision recognition, driving innovations across diverse domains ranging from media and education to healthcare and transportation. Despite its success, the robustness of multi-modal learning for visual recognition is often challenged by the unavailability of a subset of modalities, especially the visual modality. Conventional approaches to mitigate missing modalities in multi-modal learning rely heavily on algorithms and modality fusion schemes. In contrast, this paper explores the use of text-to-image models to assist multi-modal learning. Specifically, we propose a simple but effective multi-modal learning framework \texttt{GTI-MM} to enhance the data efficiency and model robustness against missing visual modality by imputing the missing data with generative transformers. Using multiple multi-modal datasets with visual recognition tasks, we present a comprehensive analysis of diverse conditions involving missing visual modality in data, including model training. Our findings reveal that synthetic images benefit training data efficiency with visual data missing in training and improve model robustness with visual data missing involving training and testing. Moreover, we demonstrate \texttt{GTI-MM} is effective with lower generation quantity and simple prompt techniques.

\end{abstract}

\input{1_intro}
\vspace{-2mm}
\input{2_related}

\vspace{-2mm}

\input{3_problem}

\vspace{-1mm}
\input{4_design}

\vspace{-2mm}
\input{5_exp}

\input{6_analysis}

\input{7_conclusion}

\bibliographystyle{plain}
\bibliography{mybib}


\appendix

\input{appendix}

\end{document}

%% file: 1_intro.tex
\section{Introduction}
\label{sec:intro}

Recent rapid advances in machine learning (ML) \cite{lecun2015deep} have revolutionized visual recognition systems and services in industries ranging from entertainment and transportation to healthcare and defense, transforming how people live, work, and interact with each other. Combining visual data with other modalities, such as audio, speech, and text, enables multi-modal systems in visual recognition to robustly perceive, interpret, and understand signals centered around human activity \cite{liang2022foundations}. Notably, researchers and AI practitioners have combined multi-modal learning with self-supervision to learn robust multi-modal representation from large-scale unlabeled datasets \cite{chen2020simple, radford2021learning}. This integration has led to the development of numerous advanced models with superior performance across a variety of visual recognition tasks, including captioning \cite{Li2022BLIPBL}, classification \cite{radford2021learning}, question-answering \cite{akbari2021vatt}, and cross-modal retrieval \cite{Kim2021ViLTVT}.

Despite the promises that multi-modal learning holds for vision recognition, this paradigm frequently hinges on the assumption that all modalities are present and available. However, in real-world settings, ensuring the completeness of modalities is challenging. Instances of missing modalities easily arise because of practical constraints such as device heterogeneity in data acquisition, hardware failures during data sampling, and privacy-related restrictions in storing, using, and sharing data \cite{voigt2017eu}. Specifically, many modalities, including videos, carry sensitive information that can reveal personally identifiable information (P.I.I.), prompting many legislations to safeguard them, including the recently introduced GDPR \cite{voigt2017eu}. Prior research \cite{Ma2021SMILML} has shown that the incomplete modality can cause substantial decreases in performance for multi-modal models.

 Even though many efforts have been made to enhance the robustness of multi-modal models against missing modalities \cite{Ma2021SMILML,ma2022multimodal,lee2023cvpr}, the effectiveness of these approaches is still constrained by the loss of information resulting from the missing modality. On the other end, generative AI is a rapidly evolving technology (e.g., ChatGPT, DALL-E-2 \cite{openai}, etc.) that enables the creation of realistic and high-fidelity digital content, such as images, based on user-input requirements or prompts. These advances in generative AI not only accelerate content generation but also present opportunities to improve the robustness of multi-modal models against missing modalities, as high-quality generated (synthetic) content offers possibilities to serve as training data. Prior works have leveraged synthetic data as training data \cite{he2022synthetic, zhang2023gpt}, reporting encouraging zero-shot performances even compared to real data. However, most of these studies focus on uni-modal setup, while there is no existing research exploring the use of synthetic data in multi-modal learning, particularly to address missing modalities.

In this work, we propose \texttt{GTI-MM}, a \textbf{G}enerative-\textbf{T}ransformer \textbf{I}mputation approach for \textbf{M}ulti-\textbf{M}odal learning that addresses the challenges caused by visual modality missing. The core concept behind \texttt{GTI-MM} is to leverage the knowledge from the generative pre-trained transformer models and to \textit{impute} missing visual data with synthetic data generation. Specifically, \texttt{GTI-MM} utilizes the generative pre-trained transformers to generate synthetic data which are then mixed with the available multi-modal training data. In order to answer whether text-to-image models can assist multi-modal learning for visual recognition with visual modality missing, we conduct a set of experiments aiming to answer the following research questions:

\begin{itemize}[leftmargin=*]
    \item Due to the privacy-sensitive nature of the visual modality, it is prevalent to train multi-modal models with missing visual data to comply with privacy regulations. \textbf{Therefore, can synthetic images improve data efficiency in multi-modal learning with visual modality missing in training?}

    \item Prior works have discovered the importance of diversity in data generation. \textbf{In this work, we would like to investigate how the quantity, complexity, and diversity of the visual imputation impact multi-modal learning with the visual modality missing.}

    \item Prior literature has proposed methods like modality dropout training and prompt learning to increase the robustness of multi-modal models against missing modalities. \textbf{Therefore, would \texttt{GTI-MM} adaptive to existing approaches in enhancing model robustness?}
    
    \item \textbf{Can \texttt{GTI-MM} generalize to different visual recognition tasks and modality missing scenarios?}
    
\end{itemize}

%% file: 2_related.tex
\section{Related Works}

\textbf{Multi-modal-learning:}
Multimodal learning \cite{liang2022foundations, Wang2019WhatMT} integrates information from diverse modalities to enable sophisticated tasks such as video-understanding \cite{monfort2019moments,wang2023internvid,maaz2023video}, text-to-image generation \cite{Rombach2021HighResolutionIS,Ruiz_2023_CVPR,Ramesh2021ZeroShotTG}, embodied navigation \cite{Paul2022AVLENAE} through various multi-scale fusion mechanisms, e.g., early, mid, and late fusion operations. With the rise in web-based data multimodal sources like LAION-5B \cite{Schuhmann2022LAION5BAO}, Conceptual-captions \cite{sharma-etal-2018-conceptual}, WIT \cite{WIT}, there has been an increasing trend toward large-scale pre-training of \cite{NIPS2017_3f5ee243} multimodal transformer models \cite{xu2023multimodal}. Multimodal transformers can be broadly classified into {dual-stream} (LXMERT \cite{Tan2019LXMERTLC}, ViLBERT \cite{Lu2019ViLBERTPT}), {single-stream} 
(Visual-BERT \cite{Li2019VisualBERTAS}, VL-BERT \cite{Su2019VLBERTPO}, ViLT \cite{Kim2021ViLTVT}, OFA \cite{Wang2022OFAUA}, MMBT \cite{Kiela2019SupervisedMB} and {encoder-decoder} (VL-T5 \cite{Cho2021UnifyingVT}, ALBEF \cite{Li2021AlignBF}, m-PLUG \cite{Xu2023mPLUG2AM}) models. 

\vspace{0.5mm}

\noindent \textbf{Missing modality:}
Prior works have handled the issue of missing modalities through incomplete sample removal \cite{HGMF,Ni2019ModelingHR} or modality imputation \cite{Zhang2022M3CareLW}. Generative methods have focused on cascaded autoencoder \cite{Tran2017MissingMI} or GAN-based \cite{Shang2017VIGANMV} approaches for learning relationships between different views/modalities in the scenario of incomplete information. Apart from generative approaches, joint learning methods have 
considered cycle-consistency \cite{Pham2018FoundIT,Zhao2021MissingMI}, and bayesian-meta-learning \cite{Ma2021SMILML} to reconstruct modalities in incomplete settings. Further, a shared-feature alignment-based approach \cite{wang2023multi} has been proposed to handle multiple tasks, including classification and segmentation across diverse domains. Multi-modal transformers have also been studied through the lens of missing modality by the inclusion of feature-reconstruction \cite{Yuan2021TransformerbasedFR}, tag-based encoding \cite{Zeng2022TagassistedMS}, and multi-task modeling \cite{ma2022multimodal}. Other extensions regarding missing modality with multi-modal transformers have been studied through the ideas of prompt learning \cite{lee2023cvpr}.

%% file: 3_problem.tex
\begin{figure}[t]
	\centering
	\includegraphics[width=0.65\linewidth]{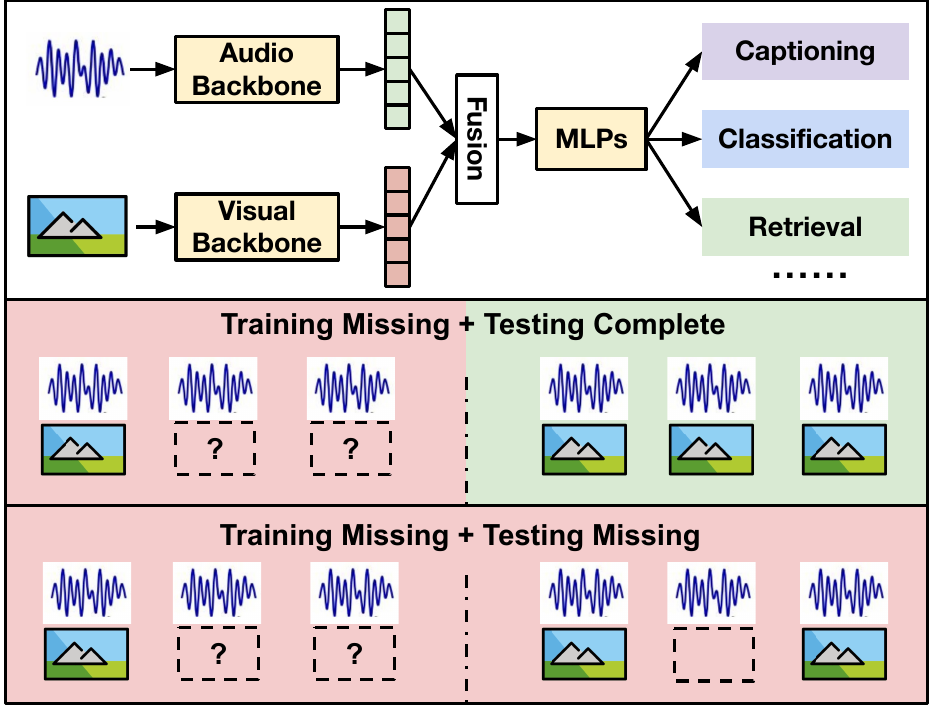}
    \caption{Problem formulation of missing modalities in this work with audio-visual recognition as the example. The missing modality includes cases in training data alone or any data.}
    \vspace{-3mm}
    \label{fig:problem_formulation}
\end{figure}

\section{Problem Formulation}
\label{sec:problem_formulation}

Our problem setup for multi-modal learning with missing modalities is shown in Figure~\ref{fig:problem_formulation}. In this work, we focus on the multi-modal learning involving audio-visual modalities. Specifically, we investigate two modality-incomplete scenarios: modality missing in training or in any data samples (both in training and testing). As mentioned earlier, \textbf{visual data is frequently associated with P.I.I. that people wish to keep private}. Therefore, \textbf{we choose visual modality as the default missing modality} in our experiments. We note that we also extend the current framework to missing audio modality in the later part of the paper.

\subsection{Visual-Modality Missing in Training Data}

To begin with, we study a relatively trivial case where missing modality occurs only in training while testing with complete modality. Similar to \cite{lee2023cvpr}, we denote our multi-modal training dataset with visual modality missing as $\mathcal{D} = \{\mathcal{D}^{C}, \mathcal{D}^{A}\}$, where $\mathcal{D}^{C}$ and $\mathcal{D}^{A}$ represents the modality-complete and audio-only data samples, respectively. More concretely, we define $\mathcal{D}^{C} = \{x_{i}^{A}, x_{i}^{V}, y_{i}\}$ and $\mathcal{D}^{A} = \{x_{i}^{A}, y_{i}\}$, where $i\in\mathbb{N}$, $A$ and $V$ represent audio and visual modalities, respectively. Moreover, we denote the visual modality missing ratio in training data as $p$. Unlike prior studies that often experiment with $p<90\%$, our investigation focuses on much more extreme settings with $p>90\%$. In this setup, the primary goal is to improve data efficiency in training.

\subsection{Visual-Modality Missing in Any Data}

In addition to missing modalities in the training data, we investigate cases where any data sample can have the visual modality missing. In this context, the testing modality loss is often regarded as an adversarial perturbation that can degrade the performance of an existing multi-modal model. Hence, increasing the model performance on modality-incomplete test data improves the model robustness.

%% file: 4_design.tex
\section{\texttt{GTI-MM} Framework}

\subsection{Multi-modal Learning Task}

\begin{figure*}[t]
	\centering
	\includegraphics[width=\linewidth]{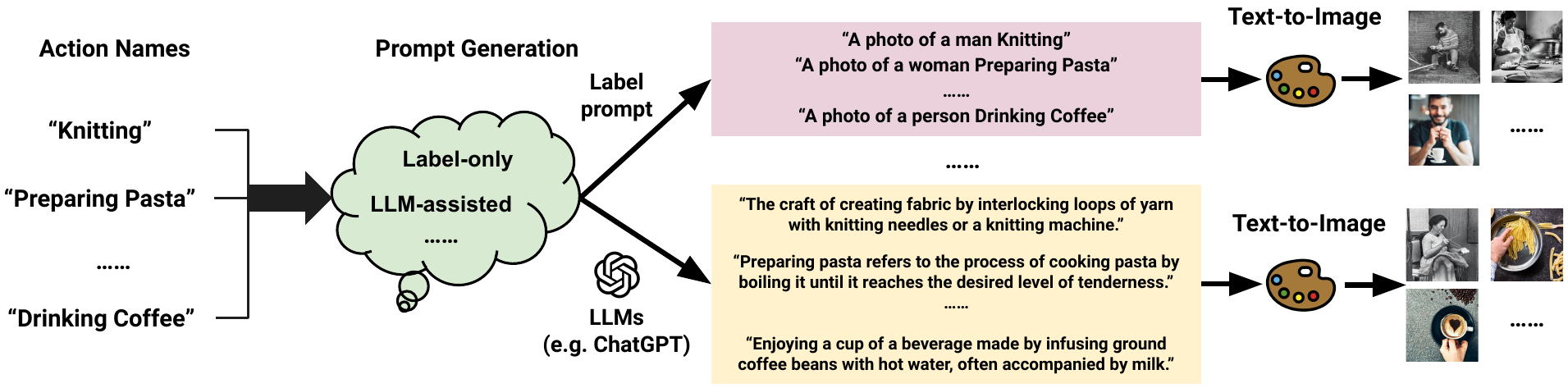}
    \caption{Visual data generation process in \texttt{GTI-MM}.}
    \label{fig:data_generation}
\end{figure*}

In this work, we explore our proposed GTI-MM framework on the multimedia human activity recognition task, which has emerged as a key focus area in large-scale video understanding. This task involves classifying a video into action categories based on underlying visual and audio modalities. This vision recognition task matches our problem formulation as most visual data consists of humans, leading to privacy concerns in leaking information like body shapes, facial geometries, and other bio-metric fingerprints.

\subsection{Pre-trained Multimodal Model}

We experiment with a recently released multi-modal model called ImageBind \cite{girdhar2023imagebind}. ImageBind proposes to learn joint multi-modal representations across 6 diverse modalities, including images, text, audio, depth, thermal, and IMU data. ImageBind leverages self-supervised learning to align images with other modalities using contrastive objectives. This model has demonstrated strong capabilities across diverse multi-modal tasks involving visual recognition. ImageBind uses Transformer architectures, where the image and audio encoder follow the Vision
Transformer \cite{dosovitskiy2020image} and Audio Spectrogram Transformer \cite{gong2021ast}, respectively.

\subsection{Visual Data Generation}

The visual data generation process used in this work is demonstrated in Figure~\ref{fig:data_generation}.
The core idea behind \texttt{GTI-MM} is to impute the missing visual modality with generative pre-trained transformers. To do so, we prompt the modern text-to-image models with the label information to generate the synthetic images. Specifically, we design our prompt messages using the following approaches:

\noindent \textbf{Label Prompt:} As demonstrated in the prior work \cite{he2022synthetic}, creating the input prompt with the class name yields visual datasets that lead to competitive zero-shot image classification performance. Therefore, we adopt this simple but effective approach to generating images related to human activities. Moreover, we augment the input prompt with different human action performers since our target application is associated with humans. Notably, given the input \texttt{ACITION\_NAME} from the set $C = \{c_{1}, ..., c_{n}\}$ and \texttt{ACTION\_PERFORMER} $s$ from the set $S =$\texttt{\{a man, a woman, a child, a person, a group of people\}}, we obtain the prompt $t$ = "\texttt{A photo of \texttt{ACTION\_PERFORMER} \texttt{ACTION\_NAME}}". As an example, for the action name "drinking coffee" and action performer "a person", \texttt{GTI-MM} uses the following prompt for the text-to-image model to generate an image: "\texttt{A photo of a person drinking coffee}." 

\noindent \textbf{LLM-assisted Prompt:} In addition to crafting prompt queries with class names, we leverage LLMs to generate prompt messages. Specifically, we adopt the following template to prompt LLMs to provide the text descriptions for the action category: \texttt{Provide 5 definitions of action class ACTION\_NAME}.

\noindent \textbf{Diversity Enhancement:} 
Prior research \cite{Shipard2023DiversityID, he2022synthetic} has demonstrated that diversity is needed in image generation for it to be used as training data. For example, \cite{Shipard2023DiversityID} experiments with enriching the diversity of the image generation by randomly setting the \textbf{unconditional guidance scale (UGC)} of the text-to-image model between 1 and 5, where a higher guidance scale indicates more creative generation. In addition to increasing the randomness of the guidance scale, researchers propose to include \textbf{multi-domain} knowledge in data generation. For example, instead of generating images belonging to "photo", we increase the diversity of the generation by augmenting the domain list to include the following: drawing, painting, sketch, collage, poster, digital art image, rock drawing, stick figure, and 3D rendering.

\begin{figure*}[t]
	\centering
	\includegraphics[width=\linewidth]{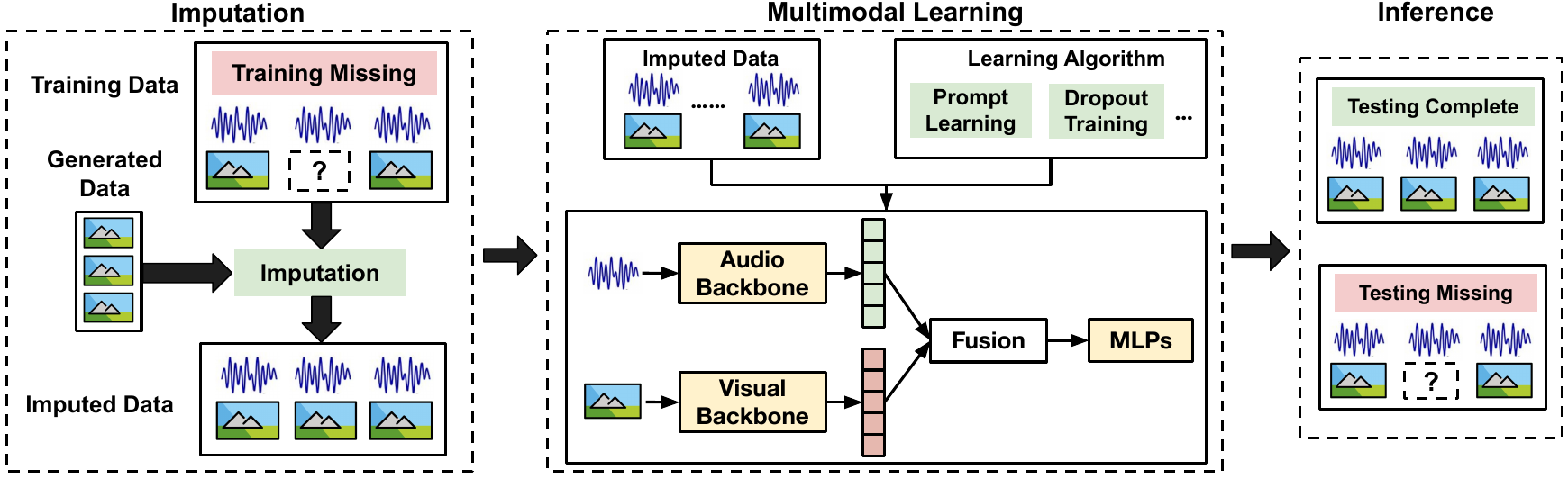}
    \caption{Learning framework of \texttt{GTI-MM}: Imputing missing visual modality with synthetic visual content for robust multi-modal learning.}
    \label{fig:multimodal_leaerning}
\end{figure*}

\subsection{Multi-modal Learning with Visual Imputation}

The multi-modal learning framework in \texttt{GTI-MM} is illustrated in Figure~\ref{fig:multimodal_leaerning}. Without loss of generality, we define the generated visual dataset as $\mathcal{D}^{V'} = \{{x}_{j}^{V'}, y_{j}\}$, where ${x}_{j}^{V'}$ is the generated image with action label $y_{j}$ and $V'$ denotes the generated visual modality. During each epoch in the multi-modal training, we impute each audio-only data by randomly selecting a generated visual sample in $\mathcal{D}^{V'}$ with the same action label. Consequently, we obtain a modality-complete dataset $\hat{\mathcal{D}} = \{\mathcal{D}^{C}, \hat{\mathcal{D}}^{A}\}$ with visual data imputation, where $\hat{\mathcal{D}}^{A} = \{x_{i}^{A}, x_{i}^{V'}, y_{i}\}$. Finally, we perform multi-modal training with the imputed dataset $\hat{\mathcal{D}}$. It is worth noting that \texttt{GTI-MM} can integrate with other multi-modal learning algorithms, such as dropout training and prompt learning, as demonstrated in Figure~\ref{fig:multimodal_leaerning}.

%% file: 5_exp.tex
\begin{table}[t]
    \caption{Summary of the audio-visual dataset statistics used in this work.}
    \centering
    \small
    \begin{tabular*}{0.55\linewidth}{lccc}
        \toprule
        
        \multirow{1}{*}{\shortstack{\textbf{Datasets}}} & 
        \multirow{1}{*}{\textbf{Video Style}} & 
        \multirow{1}{*}{\shortstack{\textbf{Classes}}} &
        \multirow{1}{*}{\shortstack{\textbf{Data Size}}}  \\ 
        \midrule
        \textbf{UCF101} & Camera & 51 & 6,837 \\ 
        \midrule
        \textbf{ActivityNet} & Camera & 200 & 18,976 \\
        \midrule
        \textbf{MiT10} & Camera, Animate & 10 & 43,460 \\
        \textbf{MiT51} & Screencast, etc. & 51 & 163,038 \\

        \bottomrule
    \end{tabular*}
    \label{table:datasets}
\end{table}

\section{Datasets and Experimental Details}

\subsection{Datasets}
This paper includes three popular multimedia action recognition datasets: UCF101 \cite{soomro2012ucf101}, ActivityNet \cite{caba2015activitynet}, and Moments in Time (MiT) \cite{monfort2019moments}. The details of the datasets are listed in Table~\ref {table:datasets}. Similar to \cite{gong2022contrastive}, we extract the video frame in the middle of the video as the visual input. Specifically, we identify that data associated with only 51 and 200 labels are presented with audio-visual data in UCF101 and ActivityNet, respectively. This leads to reduced data sizes compared to the complete data. It is worth noting that the MiT is a challenging dataset for action recognition, with SOTA accuracy close to 50\% \cite{yu2022coca,srivastava2024omnivec}. Given the inherent difficulty of the task, we tackle the easier classification problem by creating partitions of data with fewer distinct labels. Following the work in \cite{feng2023fedmultimodal}, we create 2 sub-datasets, MiT10 and MiT51, from the original MiT dataset. MiT10 and MiT51 contain videos of the 10 and the 51 most frequent labels. The details of the datasets are described in the Appendix.

\subsection{Visual Data Generation}

In this work, we use the Latent Diffusion Model~\cite{Rombach2021HighResolutionIS} loaded with Stable Diffusion V2.1 weights to generate synthetic images. Our baseline visual data generation involves the generation of 100 images of each action category associated with each dataset, leading to a total of 5,100, 20,000, 1,000, and 5,100 generated images for UCF101, ActivityNet, MiT10, and MiT51. We repeat the generation process for each prompting approach mentioned. Moreover, we choose not to impute missing data in the test set, given that the test data labels should remain unknown.

\subsection{Model Training and Evaluation}

We adopt the ImageBind Large model \cite{girdhar2023imagebind} as our visual and audio backbone. Similar to the prior work \cite{lee2023cvpr}, we froze the backbone encoders in our fine-tuning experiments, as full model fine-tuning requires substantial computation resources. We apply late fusion as our default fusion method, as we find that late fusion consistently yields the best performance across different datasets. The details of performance with other fusion approaches and advanced training approaches (e.g., OGM-GE \cite{Peng2022BalancedML}) are reported in the Appendix. We use the test accuracy of the trained model as our evaluation metric. We run the experiments with three different random seeds and report the average performance on UCF101 and ActivityNet, while we choose a fixed seed in training MiT10 and MiT51 due to the relatively larger size of the datasets. The details of the hyper-parameter selection of each dataset are described in the Appendix.

%% file: 6_analysis.tex
\section{Can \textbf{\texttt{GTI-MM}} improve data efficiency with training visual modality missing?}

\subsection{Would audio be enough for audio-visual action recognition?}

As missing visual data is present in our setup, it is natural to ask whether relying on audio data alone is adequate for action recognition. Therefore, we explore the training with complete audio data. Table~\ref{tab:baseline_results} illustrates the comparisons among audio-only, visual-only, and multi-modal models. The results reveal a substantial performance advantage with models involving the visual modality compared to relying solely on audio, implying the importance of visual data in action recognition. Furthermore, we observe that multi-modal models underperform visual-only models in some datasets. We note that this behavior may be associated with modality competition \cite{huang2022modality}, a theory suggesting that weaker modalities may be over-optimized during training, leading to decreased multi-modal performance. \textbf{Overall, using audio information alone provides limited capabilities in action recognition, and it is critical to use visual information to assist audio-visual action recognition.}

\begin{table}[t]
    \caption{Comparisons among audio-only, vision-only, and multi-modal models across different datasets. The visual missing ratio $p=0\%$ in training visual-only and multi-modal models.}
    \centering
    \small
    \begin{tabular*}{0.58\linewidth}{lccc}
        \toprule
        & 
        \multicolumn{1}{c}{\textbf{Audio-only}} &
        \multicolumn{1}{c}{\textbf{Visual-only}} & 
        \multicolumn{1}{c}{\textbf{Multi-modal}} \\

        \midrule 
        \textbf{UCF101} & $50.74$  & $\mathbf{91.08}$ & $88.75$ \\
        \textbf{ActivityNet} & $16.66$  & $\mathbf{70.72}$ & $65.66$ \\
        \textbf{MiT10} & $52.55$  & $71.48$ & $\mathbf{76.11}$ \\
        \textbf{MiT51} & $33.42$  & $58.95$ & $\mathbf{60.24}$ \\

        \bottomrule
    \end{tabular*}
\label{tab:baseline_results}
\end{table}

\subsection{Can \textbf{\texttt{GTI-MM}} improve data efficiency with severe visual modality missing in training?}

\noindent \textbf{Setup and baselines:} We investigate extreme cases of missing visual modality with low-resource training data. We set the training visual modality missing ratio $p=95\%$ for training UCF101, ActivityNet, and ActivityNet51. We apply an even larger $p=99\%$ in training MiT10 and MiT51 datasets, as these two datasets have larger data sizes than others. Our testing datasets are with complete audio and visual data. Overall, we benchmark our proposed \texttt{GTI-MM} against various baselines involving low-resource visual data:

\begin{itemize}[leftmargin=*]
    \item \textbf{Audio Training: } We explore the training with complete audio data.

    \item \textbf{Low-resource Visual Training: } This baseline is to train a uni-modal model with the available visual data. Given the missing ratio $p$, we train the model with $100-p$ visual data.

    \item \textbf{Low-resource Multi-modal Training: } One extension to low-resource visual training is adding the paired audio data. Specifically, given training missing ratio $p$, we train the model with $100-p$ paired audio-visual data.
    
    \item \textbf{Multi-modal Training with Zero-filling Imputation: } The approaches above use a portion of available data. To utilize all the available data, we impute the missing images by filling them with zeros as in \cite{lee2023cvpr}. For example, we fill $p\%$ visual data with zero in the multi-modal training.
    
\end{itemize}


\begin{table}[t]
    \caption{Performance comparisons between \texttt{GTI-MM} and other baselines across different datasets with low-resource visual data presents.}
    \centering
    \footnotesize
    \resizebox{1.0\textwidth}{!}{%
    \begin{tabular}{lcccccc}
        \toprule
        & 
        \multicolumn{1}{c}{\textbf{Training Visual}} &
        \multirow{2}{*}{\textbf{Audio Training}} &
        \multicolumn{1}{c}{\textbf{Low-source}} &
        \multicolumn{1}{c}{\textbf{Low-source}} &
        \multicolumn{1}{c}{\textbf{Zero-filling Imputation}} & 
        \textbf{\texttt{GTI-MM}} \\
        & 
        \multicolumn{1}{c}{\textbf{Missing Ratio ($p$)}} & & 
        \multicolumn{1}{c}{\textbf{Visual Training}} & 
        \multicolumn{1}{c}{\textbf{Multimodal Training}} &
        \multicolumn{1}{c}{\textbf{Multimodal Training}} & \textbf{(Ours)}\\

        \midrule 
        \textbf{UCF101} & $95\%$ & $50.74$ & $67.77$ & $32.10$ & $53.59$ & $\mathbf{78.17}$ \\
        \textbf{ActivityNet} & $95\%$ & $16.66$ & $17.10$ & $6.44$ & $17.49$ & $\mathbf{57.98}$ \\
        \textbf{MiT10} & $99\%$ & $52.55$ & $61.31$ & $48.46$ & $57.67$ & $\mathbf{68.94}$ \\
        \textbf{MiT51} & $99\%$ & $33.42$ & $38.55$ & $29.62$ & $40.18$ & $\mathbf{49.70}$ \\

        \bottomrule
    \end{tabular}
    }
\label{tab:gti_mm_baseline}
\end{table}

\vspace{-1mm}

\begin{figure*}[t] {
    \centering
    
    \begin{tikzpicture}

        \node[draw=none,fill=none] at (0,3.3){\includegraphics[width=0.45\linewidth]{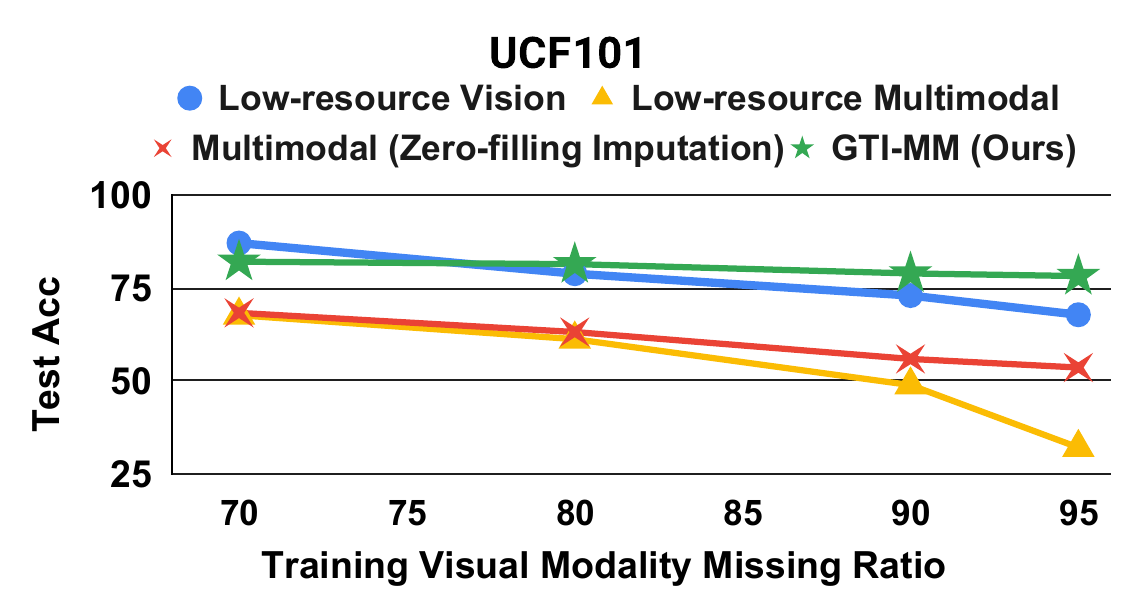}};

        \node[draw=none,fill=none] at (0.5\linewidth,3.3){\includegraphics[width=0.45\linewidth]{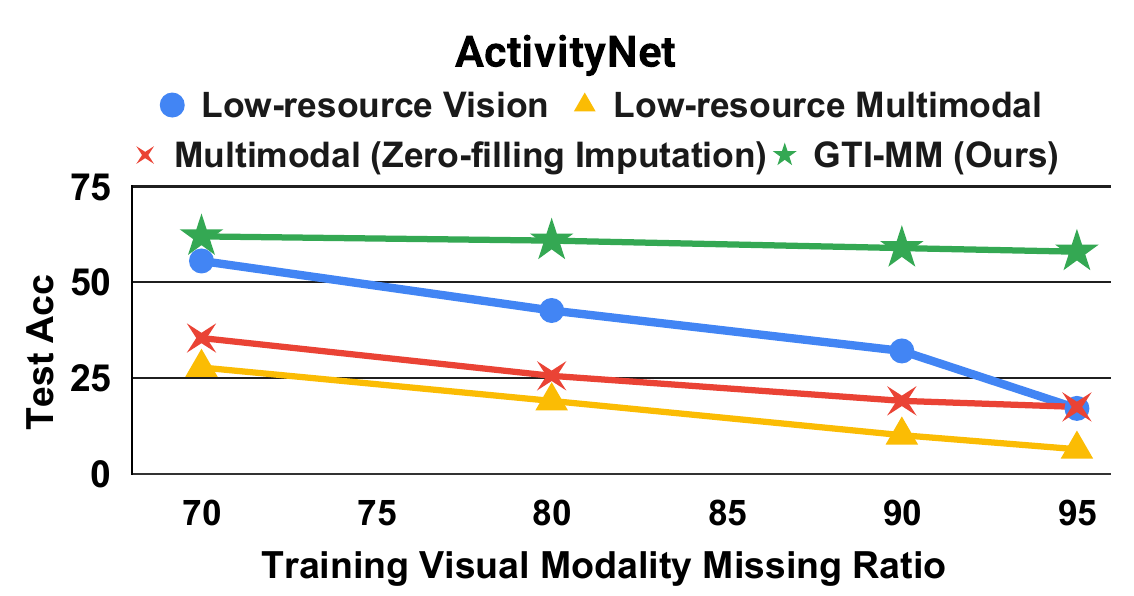}};

        \node[draw=none,fill=none] at (0.25\linewidth,0){\includegraphics[width=0.45\linewidth]{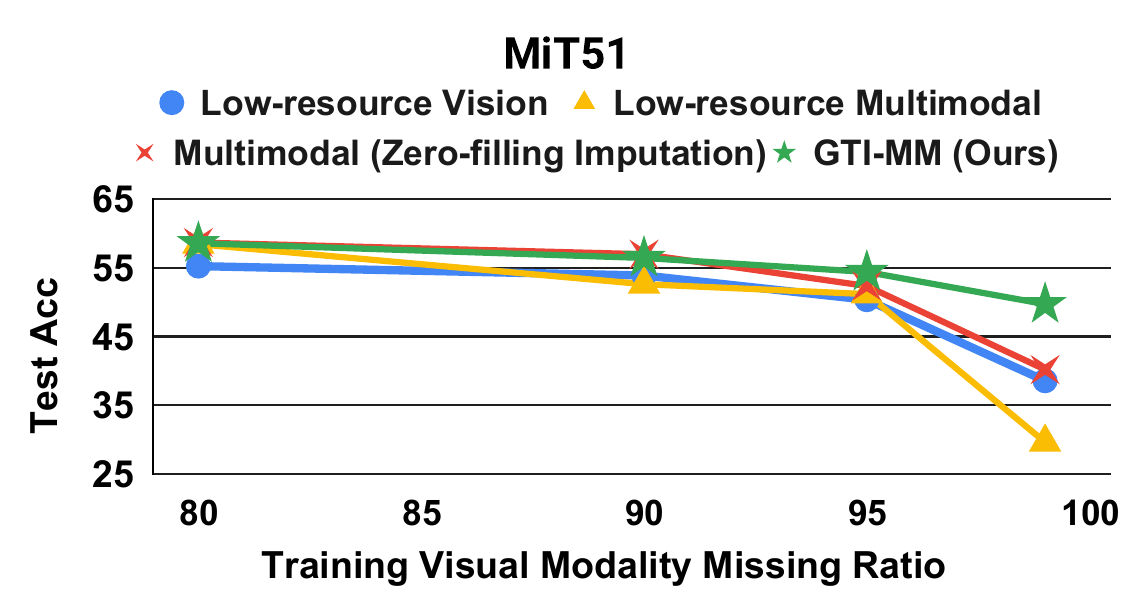}};
        
    \end{tikzpicture}
    \caption{Performance comparisons among \texttt{GTI-MM} and other baselines at different training visual modality ratios.}
    \label{fig:missing_ratio}
    \vspace{-5.5mm}
} \end{figure*}

Table~\ref{tab:gti_mm_baseline} presents the performance comparisons between \texttt{GTI-MM} and other baseline models. In this experiment, GTI-MM employs the label prompt to generate visual data, producing 100 images per class category. The results show that low-source visual training consistently yields the best performance among all baseline methods except for MiT51 datasets. In addition, we identify that modality-competition increases as low-resource multi-modal training consistently yields much worse performance than other baselines. On the other hand, we can find that multi-modal training with zero-filling imputation can further improve the model performance compared to low-source visual training. Encouragingly, the results demonstrate that our proposed multi-modal learning approach \texttt{GTI-MM} can substantially improve the multi-modal models from all existing baselines across datasets by imputing missing visual data with synthetic images. Notably, we even observe a 40\% performance increase on the ActivityNet dataset. \textbf{In summary, the results indicate that synthetic images are effective in improving data efficiency with visual data missing training.}

\begin{figure}[t]
	\centering
	\includegraphics[width=0.65\linewidth]{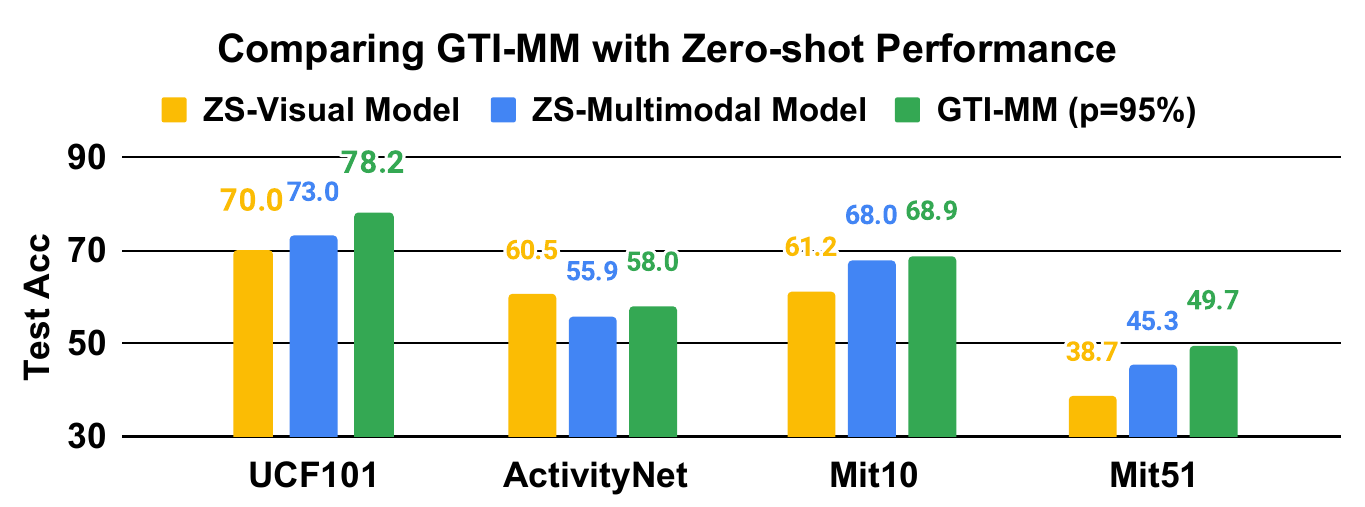}
    \caption{Comparisons between \texttt{GTI-MM} and zero-shot learning with synthetic visual data.}
    \label{fig:zs_performance}
\end{figure}

\begin{figure}[t]
	\centering
	\includegraphics[width=0.65\linewidth]{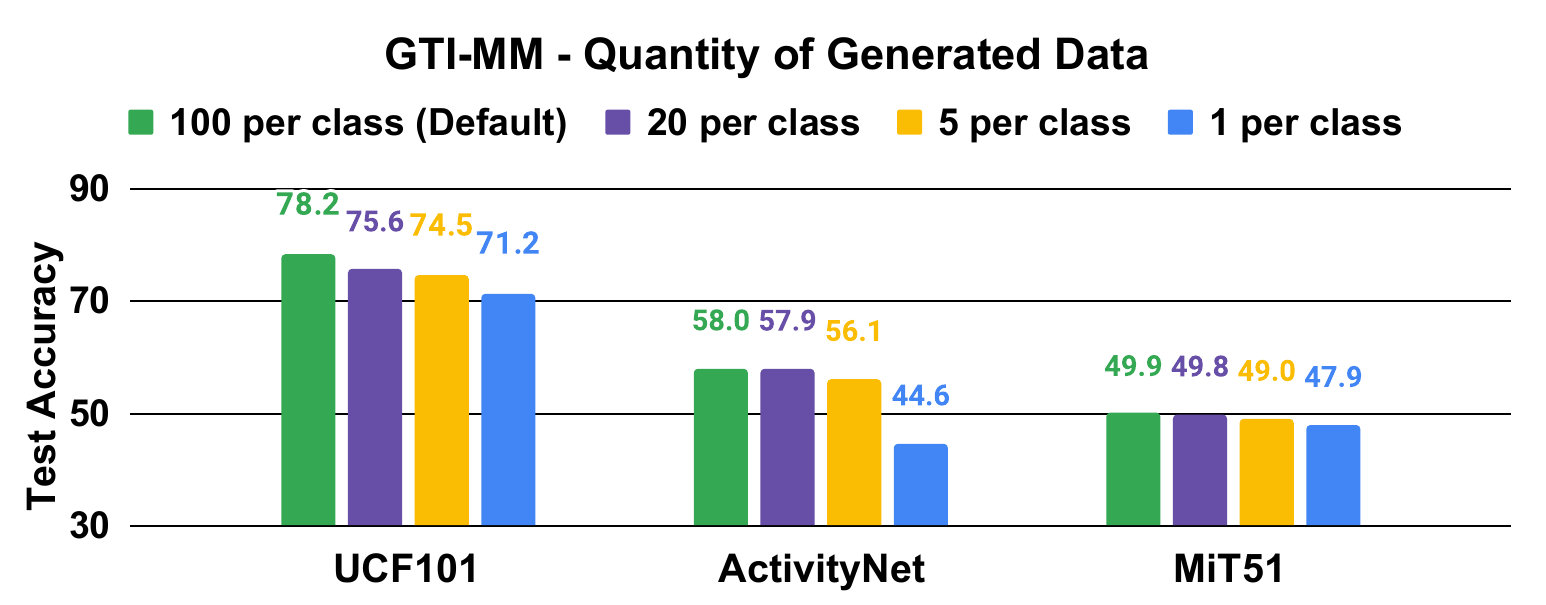}
    \caption{Impact of generation quantity on \texttt{GTI-MM} performance.}
    \label{fig:quantity}
\end{figure}

\vspace{-1mm}

\subsection{Can \textbf{\texttt{GTI-MM}} improve data efficiency with visual data missing in training with lower $p$?}
\label{subsec:comparing_p}

As shown in Table~\ref{tab:baseline_results}, in comparison to the model performance when the visual modality is complete, \texttt{GTI-MM} still exhibits lower performance. This implies that training with available visual modality would likely outperform \texttt{GTI-MM} as the training missing ratio decreases. To verify this assumption, we plot the performance of \texttt{GTI-MM} at different visual-modality missing ratios in training, as demonstrated in Figure~\ref{fig:missing_ratio}. The plot shows that as the visual modality missing ratio decreases, the performance differences between the low-resource visual model and \texttt{GTI-MM} are reasonably close, with the low-resource visual model starting to outperform \texttt{GTI-MM} in the UCF101 dataset with $p=70\%$.

\vspace{-1mm}
\subsection{Zero-shot capability with \texttt{GTI-MM}}

Table~\ref{tab:gti_mm_baseline} shows that imputing missing visual data with synthetic images can outperform all baseline approaches by a large margin, it is reasonable to hypothesize that synthetic data alone can provide competitive performance without audio data or low-source visual data in the action recognition task. Here, we propose to compare \texttt{GTI-MM} against two zero-shot learning baselines with synthetic data:

\begin{itemize}[leftmargin=*]
    \item \textbf{ZS-Visual Model} is a visual model trained with only synthetic images.

    \item \textbf{ZS-Multi-modal Model} is multi-modal learning with audio data paired with random synthetic images of the same label. 

\end{itemize}

Figure~\ref{fig:zs_performance} plots the comparisons between \texttt{GTI-MM} and zero-shot learning baselines. The comparisons between the ZS-Visual and the ZS-Multi-modal models suggest that adding audio modality provides advantages for improving model performance across most datasets. Moreover, leveraging \texttt{GTI-MM}, which involves training with low-resource visual data, demonstrates a further performance increase compared to ZS-Multi-modal and ZS-Visual models, except for the ActivityNet. Specifically, \texttt{GTI-MM} outperforms the ZS-Visual model by more than 10\% in the MiT51 dataset. These comparisons imply the need to include the weaker modality and low-resource visual data in \texttt{GTI-MM}.

\begin{figure}[t]
	\centering

    \begin{tikzpicture}

        \node[draw=none,fill=none] at (0,0){\includegraphics[width=0.5\linewidth]{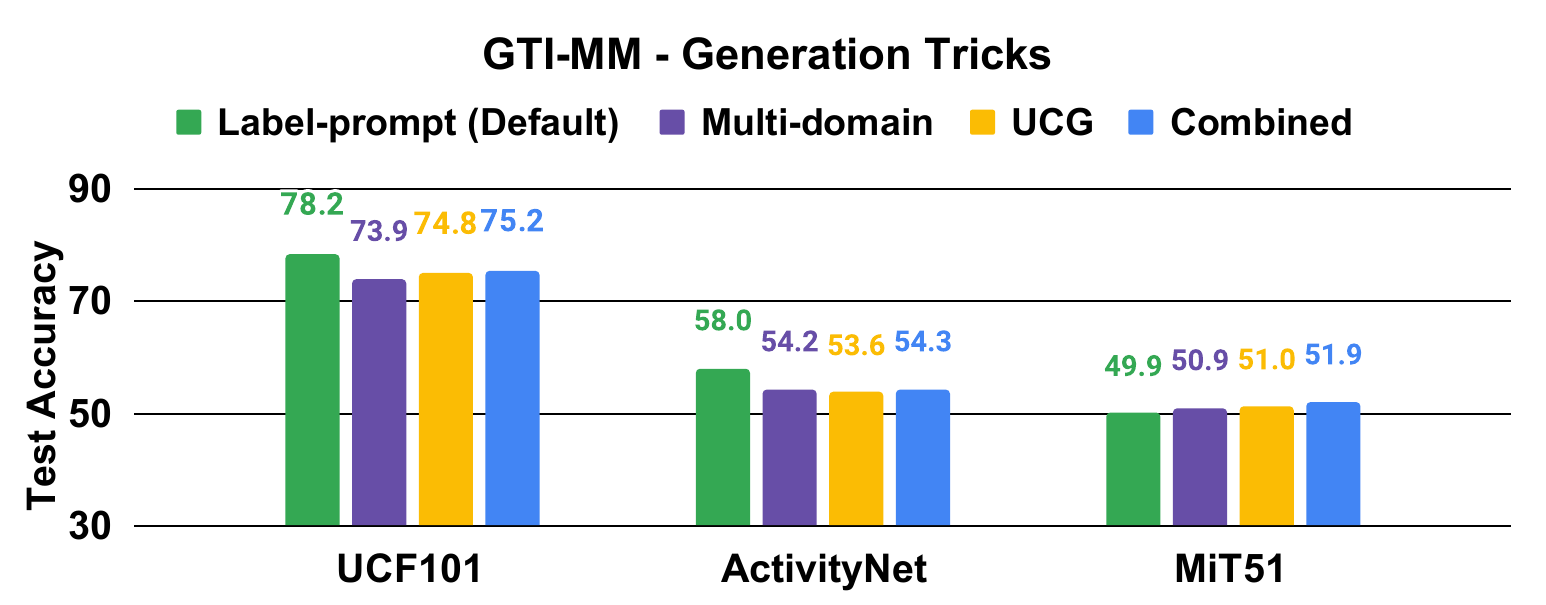}};

        \node[draw=none,fill=none] at (0.5\linewidth,0){\includegraphics[width=0.5\linewidth]{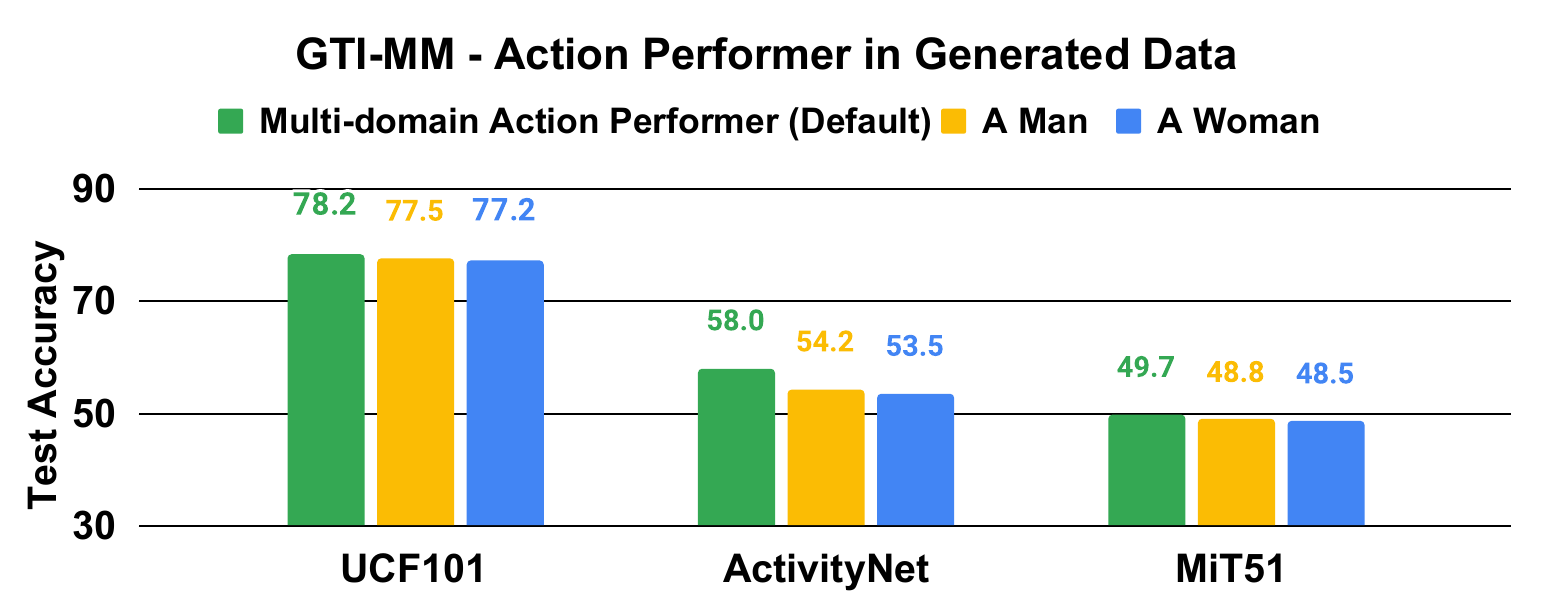}};
        
    \end{tikzpicture}
    
    \caption{Impact of generation diversity on \texttt{GTI-MM} performance.}
    \label{fig:tricks}
\end{figure}

\section{How would the quantity, complexity, and diversity in visual imputation impact multi-modal learning with training visual modality missing?}

\vspace{-1mm}
\subsection{Quantity of Visual Generation}

We identified that image generation is costly, and generating one image on an A40 GPU takes approximately 12 seconds. This prompts the need to study the quantity of generation required for \texttt{GTI-MM} to achieve a competitive performance. To do so, we perform \texttt{GTI-MM} varying number of generations per class in $\{100, 20, 5, 1\}$. Figure~\ref{fig:quantity} shows the performance of \texttt{GTI-MM} at different numbers of generations. The results are with 3 representative datasets due to limited space, and the remaining results are in the Appendix. The results indicate that 20 generations per class can provide acceptable performance compared to 100 unique generations. However, the performance of \texttt{GTI-MM} starts to drop substantially with less than 5 unique generations.

\vspace{-1mm}
\subsection{Diversity of Visual Generation}

As we discussed earlier, increasing the diversity of the data generation can lead to a positive impact on the model performance with synthetic data. Here, we investigate the impact of generation diversity on \texttt{GTI-MM} in two scenarios.

\noindent \textbf{Generation Tricks:} We compare different prompt tricks in \texttt{GTI-MM} as shown in Figure~\ref{fig:tricks}. These prompt tricks include adding multi-domain information, incorporating UGC, and combining both. \textbf{Interestingly, adding generation tricks does not guarantee improved performance}, where performances drop in ActivityNet and UCF101. However, we observe an increase in performance on the MiT51. One plausible explanation for this result is associated with the video styles originating from the dataset, where UCF101 and ActivityNet include only camera recordings, while the MiT contains videos from diverse domains, as demonstrated in Table~\ref{table:datasets}.

\begin{table}[t]
    \caption{Performance comparisons between label prompt and LLM-assist prompt in image generation.}
    \centering
    \footnotesize
    \begin{tabular*}{0.67\linewidth}{lccccc}
        \toprule
        & 
        \textbf{UCF101} &
        \textbf{ActivityNet} & 
        \textbf{MiT10} & 
        \multicolumn{1}{c}{\textbf{MiT51}} \\

        \midrule 
        \textbf{Label prompt} & $78.17$  & $57.98$ & $\mathbf{68.94}$ & $49.87$ \\

        \textbf{LLM-assist prompt} & $\mathbf{78.42}$  & $\mathbf{58.78}$ & $68.04$ & $\mathbf{50.23}$ \\
        
        \hline
    \end{tabular*}
    \label{tab:prompt_template}
\end{table}

\noindent \textbf{Action Performers:} Instead of prompting the text-to-image models to generate images with different action performers, we decided to restrict the action performer to "a man" or "a woman". Figure~\ref{fig:action_performer} shows the performance of \texttt{GTI-MM} varying action performers. \textbf{From the plot, it is evident that incorporating multi-domain information to action performers enhances the model performance.} This finding reveals the importance of guiding future research in studying fairness challenges with training synthetic data. 

\subsection{Complexity in Prompts}
Table~\ref{tab:prompt_template} compares the image generation between employing label prompt and LLM-assist prompt. Specifically, we instructed ChatGPT to provide 5 descriptions of each activity. The results indicate that supplementing the prompt message with a concise number of detailed descriptions using LLMs can improve the performance of \texttt{GTI-MM} in the majority of datasets. \textbf{However, the performance gain with the LLM-assist prompt is marginal, indicating that the label prompt is adequate if LLMs are unavailable.}

\begin{table}[t]
    \caption{Comparisons between MM-Dropout and \texttt{GTI-MM Dropout}. $p$ and $q$ are training and testing visual missing ratios.}
    \centering
    \footnotesize
    \begin{tabular*}{0.65\linewidth}{cccccc}
        \toprule
        \multirow{2}{*}{\textbf{Dataset}} & 
        \multirow{2}{*}{\textbf{Method}} &
        \multirow{2}{*}{$\mathbf{p}$} &
        \multicolumn{3}{c}{\textbf{Test Missing Ratio ($q$)}} \\
        
        & & &
        \multicolumn{1}{c}{\textbf{50}} &
        \multicolumn{1}{c}{\textbf{70}} &
        \multicolumn{1}{c}{\textbf{90}} \\

        \cmidrule(lr){1-6}
        \multirow{3}{*}{\textbf{UCF101}} & 
        \textbf{MM-Dropout} & 
        $0$ & 
        $\mathbf{62.3}$ & 
        $56.9$ & 
        $51.7$ \\

        \cline{2-2}
        & 
        \multirow{2}{*}{\textbf{GTI-MM Dropout}} &
        $90$ & 
        $60.8$ & 
        $\mathbf{57.7}$ & 
        $\mathbf{53.8}$ \\
        
        & 
         &
        $95$ & 
        $59.4$ & 
        $56.8$ & 
        $52.5$ \\

        \cmidrule(lr){1-6}

        \multirow{3}{*}{\textbf{ActivityNet}} & 
        \textbf{MM-Dropout} & 
        $0$ & 
        $\mathbf{33.2}$ & 
        $22.9$ & 
        $17.8$ \\

        \cline{2-2}
        & 
        \multirow{2}{*}{\textbf{GTI-MM Dropout}} & $90$ & $32.7$ & $\mathbf{26.4}$ & $\mathbf{20.3}$ \\

        & & $95$ & $31.7$ & $26.0$ & $20.1$ \\

        \cmidrule(lr){1-6}

        \multirow{3}{*}{\textbf{MiT10}} & \textbf{MM-Dropout} & $0$ & $\mathbf{63.9}$ & $\mathbf{58.1}$ & $53.5$ \\

        \cline{2-2}
        & \multirow{2}{*}{\textbf{GTI-MM Dropout}} & $90$ & $60.6$ &  $57.9$ &  $\mathbf{54.7}$\\

        & & $99$ & $59.7$ & $56.4$ & $54.0$ \\

        \cmidrule(lr){1-6}

        \multirow{2}{*}{\textbf{MiT51}} & \textbf{MM-Dropout} & $0$ & $\mathbf{46.1}$ & $\mathbf{41.5}$ & $\mathbf{35.7}$ \\

        \cline{2-2}
        & \multirow{2}{*}{\textbf{GTI-MM Dropout}} & $90$ & $43.2$ & $39.3$ & $34.2$ \\

        & & $99$ & $40.3$ & $37.7$ & $34.1$ \\
        
        \bottomrule
    \end{tabular*}
    \label{tab:dropout_gti_mm}
\end{table}

\begin{figure*}[t] {
    \centering
    
    \begin{tikzpicture}

        \node[draw=none,fill=none] at (0,0){\includegraphics[width=0.335\linewidth]{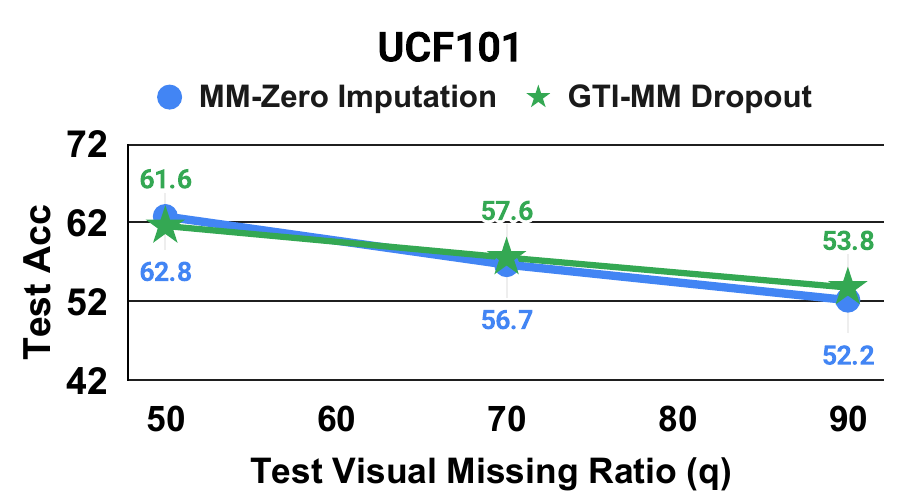}};

        \node[draw=none,fill=none] at (0.33\linewidth,0){\includegraphics[width=0.335\linewidth]{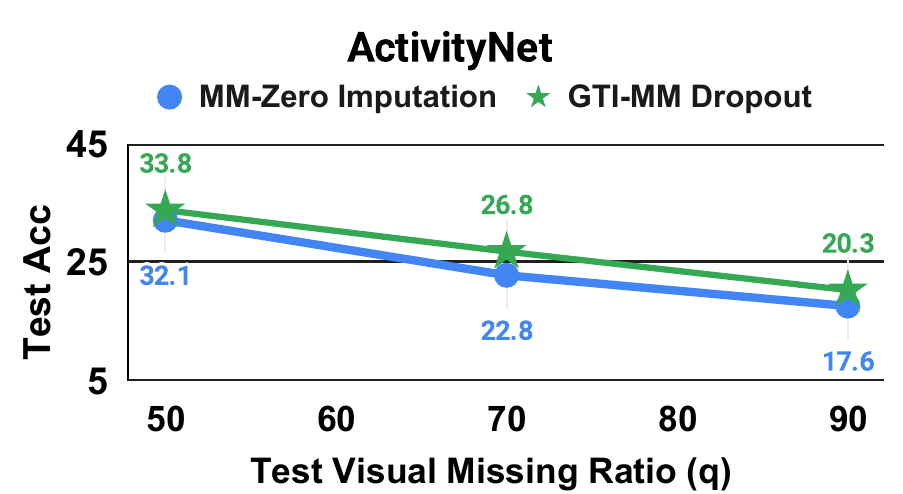}};

        \node[draw=none,fill=none] at (0.66\linewidth,0){\includegraphics[width=0.335\linewidth]{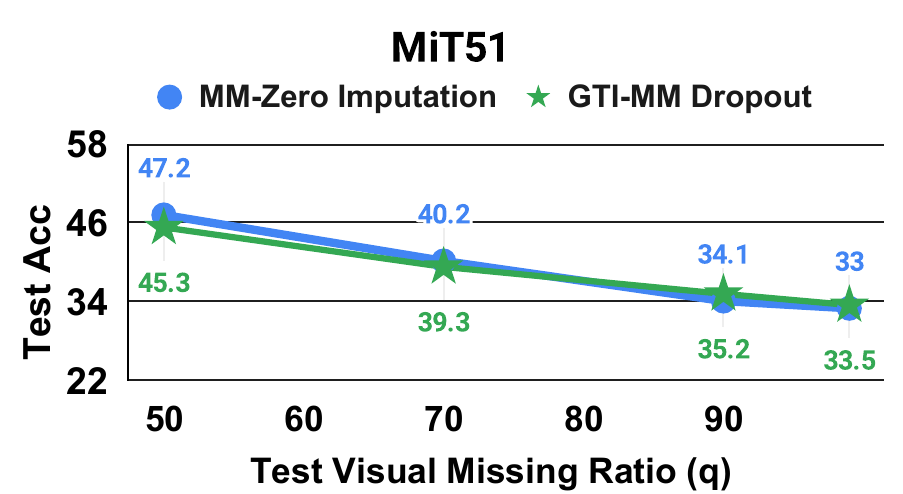}};
        
    \end{tikzpicture}
    \caption{Comparisons on testing visual modality missing between MM-Zero Imputation and \texttt{GTI-MM Dropout}, where $p=q$.}
    \label{fig:dropout_dt}
} \end{figure*}

\section{Can \texttt{GTI-MM} improve model robustness with visual modality missing?}
\label{sec:mm_any_data}

This section investigates a more severe condition where the missing visual modality can occur in both training and testing data. Dropout training \cite{ma2022multimodal} is widely applied to enhance the robustness of the multimodal models against missing modalities during inference. In this setup, we decided only to drop the visual modality during training. We evaluate with testing missing ratio $q=\{50\%, 70\%, 90\%\}$. We propose the following baselines:

\begin{itemize}[leftmargin=*]
    \item \textbf{MM-Dropout} is multimodal dropout training with complete data and serves as a strong baseline, given that it uses a complete training dataset. The dropout training is with the dropout rate equal to the testing visual missing ratio $q$.

    \item \textbf{MM-Zero Imputation} involves training a multimodal model by imputing missing visual training data with zeros, a baseline for assessing the robustness of the multimodal model used in \cite{lee2023cvpr}. We evaluate scenarios where $p=q$.

\end{itemize}

\vspace{-1mm}
\subsection{Can \textbf{\texttt{GTI-MM}} improve model robustness using Dropout Training?}

In this experiment, we combine the dropout training with \texttt{GTI-MM}. In the dropout training, we randomly fill the data samples in the imputed dataset $\hat{\mathcal{D}} = \{\mathcal{D}^{C}, \hat{\mathcal{D}}^{A}\}$ with zero. 

\vspace{0.5mm}

\noindent \textbf{Can \texttt{GTI-MM Dropout} outperform MM-Dropout?} Here, we train \texttt{GTI-MM} with the same $p$ as used in the last section, meaning that only 1\% or 5\% of visual data is available. In addition, we experiment \texttt{GTI-MM} with a relatively lower $p$ at $90\%$. The comparisons between MM-Dropout and \texttt{GTI-MM Dropout} are listed in Table~\ref{tab:dropout_gti_mm}. The results show that \texttt{GTI-MM} with dropout training is more robust than the multi-modal dropout training with complete modalities when $q$ is above 70\% even when $p\geq95\%$. However, \texttt{GTI-MM} with dropout training yields worse performances when $q=50\%$. This performance gap may be related to the fact that only $5\%$ or $1\%$ of visual data is available in \texttt{GTI-MM}. At the same time, MM-Dropout utilizes $50\%$ of the complete visual data in each training epoch. 

\vspace{0.5mm}
\noindent \textbf{Is \texttt{GTI-MM Dropout} effective when $\mathbf{p=q}$?} Here, we compare \texttt{GTI-MM Dropout} with MM-Zero Imputation by controlling $p=q$. Our results in Figure~\ref{fig:dropout_dt} reveal that \texttt{GTI-MM Dropout} provides substantial advantages in improving the performance compared to MM-Zero imputation in UCF101 and ActivityNet and marginal performance gain in MiT51 with a larger $p$, demonstrating the effectiveness of \texttt{GTI-MM} in different visual missing scenarios.

\vspace{-1mm}
\subsection{Can \textbf{\texttt{GTI-MM}} improve model robustness using prompt learning?}

As discussed earlier, missing-aware prompt learning \cite{lee2023cvpr} is one state-of-the-art approach to enhance the model robustness in testing modalities missing. Here, we extend \texttt{GTI-MM} dropout training with missing-aware prompts. Owing to the large size of the ImageBind model, we choose to insert learnable prompts only in the last layer of image and audio encoders. We set the number of learnable tokens as 5. The details of modality missing-aware prompt learning are presented in the Appendix. Table~\ref{tab:prompt_learning} compares the performance of \texttt{GTI-MM Dropout} with and without prompt learning. The results show that combining prompt learning with \texttt{GTI-MM Dropout} training can consistently increase the model performance against visual modality missing in test data. This encouraging finding supports that our proposed \texttt{GTI-MM} generalizes well across other SOTA algorithms in enhancing model performance against missing visual modality.

\begin{table}[t]
    \caption{Comparing \texttt{GTI-MM DT} with and without prompt learning on the condition where visual modality is missing in test data. $p=99\%$ for MiT datasets, while $p=95\%$ for other datasets.}
    \centering
    \footnotesize
    \begin{tabular*}{0.67\linewidth}{ccccccc}
        \toprule
        \multirow{2}{*}{\textbf{Dataset}} & 
        \multirow{2}{*}{\textbf{Prompt Learning}} &
        \multirow{2}{*}{$\mathbf{p(\%)}$} &
        \multicolumn{3}{c}{\textbf{Test Missing Ratio ($q$)}} \\
        
        & & &
        \multicolumn{1}{c}{\textbf{50}} &
        \multicolumn{1}{c}{\textbf{70}} &
        \multicolumn{1}{c}{\textbf{90}} \\

        \cmidrule(lr){1-6}
        
        \multirow{2}{*}{\textbf{UCF101}} & 
        \xmark & 
        \multirow{2}{*}{$95$} & 
        $59.4$ & 
        $56.8$ & 
        $52.5$ \\

        &
        \cmark & &
        $\mathbf{60.7}$ & 
        $\mathbf{57.3}$ &  
        $\mathbf{53.7}$ \\

        \cmidrule(lr){1-6}
        \multirow{2}{*}{\textbf{ActivityNet}} & 
        \xmark & 
        \multirow{2}{*}{$95$} & 
        $31.7$ & 
        $26.0$ & 
        $20.1$ \\

        &
        \cmark & &
        $\mathbf{32.2}$ & 
        $\mathbf{26.8}$ & 
        $\mathbf{20.6}$ \\

        \cmidrule(lr){1-6}
        \multirow{2}{*}{\textbf{MiT10}} & 
        \xmark & 
        \multirow{2}{*}{$99$} & 
        $59.7$ & 
        $56.4$ & 
        $54.0$ \\

        &
        \cmark & &
        $\mathbf{60.8}$ & 
        $\mathbf{55.7}$ & 
        $\mathbf{54.8}$ \\

        \cmidrule(lr){1-6}
        \multirow{2}{*}{\textbf{MiT51}} & 
        \xmark & 
        \multirow{2}{*}{$99$} & 
        $40.3$ &
        $37.7$ & 
        $34.1$ \\

        &
        \cmark & &
        $\mathbf{42.1}$ & 
        $\mathbf{37.7}$ & 
        $\mathbf{34.4}$  \\
        
        \bottomrule
    \end{tabular*}
    \label{tab:prompt_learning}
\end{table}

\section{Generalizability of \textbf{\texttt{GTI-MM}}}

\subsection{Can \textbf{\texttt{GTI-MM}} adapt to text-visual learning tasks?}
As our experiments focus on audio-visual datasets for activity recognition, there is uncertainty about the effectiveness of \texttt{GTI-MM} in multi-modal tasks involving different modalities. To address this concern, we perform additional experiments on the text-visual classification. Our experiments demonstrate that \texttt{GTI-MM} is adaptable to diverse multi-modal tasks involving text-visual data. The details about the text-visual experiments are described in the Appendix.

\subsection{Can we extend visual imputation to audio imputation with audio modality missing?}
Our comprehensive results indicate that \texttt{GTI-MM} is capable of improving data efficiency and model robustness against visual modality missing. However, the effectiveness of \texttt{GTI-MM} in scenarios where the audio is missing remains uncertain. To answer this, we explore the audio data imputation within the \texttt{GTI-MM}, utilizing the AudioLDM2 \cite{liu2023audioldm} for generating audio data. In contrast to visual data imputation, results in Table~\ref{tab:audio_data} reveal that \texttt{GTI-MM} underperforms zero-filling imputation in most datasets. This suggests that existing audio generation models may struggle to generate adequate-quality audio, underscoring the need to develop more advanced audio generation models. However, we identify that audio imputation still benefits model performance in ActivityNet, suggesting the potential of \texttt{GTI-MM} in audio imputation. The details about the audio imputation experiments are included in the Appendix. Given the capabilities of advanced LLMs like ChatGPT \cite{openai}, we choose to study audio imputation as audio generation is more challenging than text generation.

\begin{table}[t]
    \caption{Comparisons between audio imputation with zero-filling and \texttt{GTI-MM} in the case of training audio modality missing. $p$ is the same as in training visual missing experiments.}
    \centering
    \footnotesize
    \begin{tabular*}{0.6\linewidth}{lccccc}
        \toprule
        & 
        \textbf{UCF101} &
        \textbf{ActivityNet} &
        \textbf{MiT10} &
        \textbf{MiT51} \\

        \midrule 
        \textbf{Zero-Filling} & $\mathbf{87.0}$  & $45.6$ & $\mathbf{70.87}$ & $\mathbf{55.5}$ \\

        \textbf{\texttt{GTI-MM}} & $79.9$ & $\mathbf{54.4}$ & $60.75$ & $47.2$ \\
        
        \bottomrule
    \end{tabular*}
    \label{tab:audio_data}
\end{table}

%% file: 7_conclusion.tex
\section{Conclusions}

We proposed \texttt{GTI-MM}, a generative-transformer imputation approach, for multi-modal learning to address the challenges caused by missing visual modality. Our extensive experiments demonstrate that \texttt{GTI-MM} provides robust multi-modal solutions against severe missing visual modality settings in training data or testing data. Crucially, increasing the diversity, quantity, and complexity of the prompting approach further enhances \texttt{GTI-MM} performance. While \texttt{GTI-MM} is effective in imputing visual data, it encounters challenges in audio data imputation. Our future work plans to explore more sophisticated generative transformers to further improve the quality of the data generation, such as models with multi-modal capabilities (e.g., CoDi \cite{tang2023any}).

%% file: appendix.tex
\newpage
\section{Appendix} \label{Appendix}

\section{Modality Missing Aware Prompt Learning}
In this section, we provide brief descriptions of the missing-aware prompt learning used in this work. The details about the approach can be referenced in \cite{lee2023cvpr}. Here, we define the input embeddings to $i$-th layer of the visual encoder and audio encoder as $h^{v}_{i} \in \mathcal{R}^{L^{v} \times d^{v}}$ and $h^{a}_{i} \in \mathcal{R}^{L^{a} \times d^{a}}$, respectively. Specifically, $L$ and $d$ represent the length of the input and the dimension of the input embedding. Then, we define the missing-aware prompt as $p^{v}_{i} \in \mathcal{R}^{L \times d^{v}}$ and $p^{a}_{i} \in \mathcal{R}^{L \times d^{a}}$ for the visual and audio modality, respectively, where $L$ is the prompt length. Finally, we insert these missing-aware prompts at the beginning of the input embeddings. For example, the resulting input to the $i$-th layer of the visual encoder is: $h^{v}_{i, prompt} = (p^{v}_{i}, h^{v}_{i}) \in \mathcal{R}^{(L^{v}+L) \times d^{v}}$. Due to the large parameter size of the ImageBind model, we only insert the prompts at the last layer of the visual and audio encoders. During training, only the inserted prompts are being updated while the multi-modal transformer parameters are frozen at all times. We want to highlight that our proposed \texttt{GTI-MM} framework is not only adaptable to the modality missing aware prompt learning but also other multi-modal learning algorithms.

\section{Dataset Details}
\label{sec:dataset}

\noindent \textbf{UCF101} dataset \cite{soomro2012ucf101} includes 13,320 videos with 101 sport-based action labels collected from web sources. During our data preprocessing, we identified that data associated with only 51 labels are presented with video and audio modalities, reducing total videos below 7,000. We truncate the audio data to 5 seconds during training and testing, as training audio data for ImageBind are short audio clips. 

\vspace{1mm}
\noindent \textbf{ActivityNet} \cite{caba2015activitynet} is designed to study human activity understanding, covering 203 diverse human activities in our daily lives. Similar to UCF101, we identified only 200 action categories with audio and visual modalities, leading to 18,976 data instances. 

\vspace{1mm}
\noindent \textbf{Moments in Time (MiT)} is a large-scale multimedia activity recognition (approximately 1 million) dataset \cite{monfort2019moments} with short (3 seconds) videos with an overall list of 339 action labels. Unlike UCF101 and ActivityNet datasets, where the majority of the videos are camera recordings, the MiT dataset consists of video clips in diverse styles, including animated, screencast, and montage. It is worth noting that the MiT is a challenging dataset, with SOTA accuracy close to 35\% \cite{ryoo2019assemblenet}. Given the inherent difficulty of this task, we tackle the easier classification problem by creating partitions of data with fewer distinct labels. Following the prior work presented in \cite{feng2023fedmultimodal}, we create two sub-datasets, MiT10 and MiT51, from the original MiT dataset. MiT10 and MiT51 contain videos of the 10 and the 51 most frequent labels. 

\vspace{1mm}
\noindent \textbf{CrisisMMD} \cite{alam2018crisismmd} contains 18.1k tweets involving paired visual and textual data. The researcher collected relevant tweets from seven prominent natural disasters, such as Hurricane Harvey (2017). One major objective of the dataset is to identify the humanitarian aspects of the disaster, such as infrastructure damage and rescue efforts. We exclude the labels "Not humanitarian" and "Other relevant information" due to the ambiguous nature associated with the annotation.

\begin{table}[t]
    \centering
    \caption{Experimental setup of \texttt{GTI-MM} on audio-visual datasets.}
    \label{tab:av_experiments}
    \resizebox{0.65\linewidth}{!}{
        \begin{tabular}{lccccc} 
            \toprule
            \multicolumn{1}{c}{} & \textbf{UCF101} & \textbf{ActivityNet} & \textbf{MiT10} & \textbf{MiT51}  \\ 
            \cmidrule{1-5}
            Total Epoch & 15 & 15 & 15 & 15 \\
            Batch Size & 128 & 128 & 128 & 128 \\
            {Learning Rate~} & 5.00E-04 & 5.00E-04 & 5.00E-04 & 5.00E-04 \\
            Optimizer & AdamW & AdamW & AdamW & AdamW \\
            Weight Decay & 1.00E-04 & 1.00E-04 & 1.00E-04 & 1.00E-04 \\
            \bottomrule
        \end{tabular}
    }
    \label{hyper:av_experiments}
\end{table}

\begin{table}[t]
    \centering
    \caption{Experimental setup of \texttt{GTI-MM} on text-visual datasets.}
    \label{tab:tv_experiments}
    \small
    \resizebox{0.5\linewidth}{!}{
        \begin{tabular}{lcccc} 
            \toprule
            \multicolumn{1}{c}{} & \textbf{UPMC-Food101} & \textbf{Crisis-MMD} \\ 
            \cmidrule{1-3}
            Total Epoch & 30 & 30 \\
            Batch Size & 128 & 128 \\
            {Learning Rate~} & 5.00E-04 & 5.00E-04 \\
            Optimizer & AdamW & AdamW \\
            Weight Decay & 1.00E-04 & 1.00E-04 \\
            \bottomrule
        \end{tabular}
    }
    \label{hyper:tv_experiments}
\end{table}

\vspace{1mm}
\noindent \textbf{UPMC Food101} \cite{wang2015recipe} database comprises web pages with textual recipe descriptions for 101 food categories retrieved from online sources. Each page was paired with a single image, where the images were obtained by querying Google Image Search for the given label. The web pages were processed with html2text to obtain the raw text.

\begin{figure*}[t]
	\centering
	\includegraphics[width=0.9\linewidth]{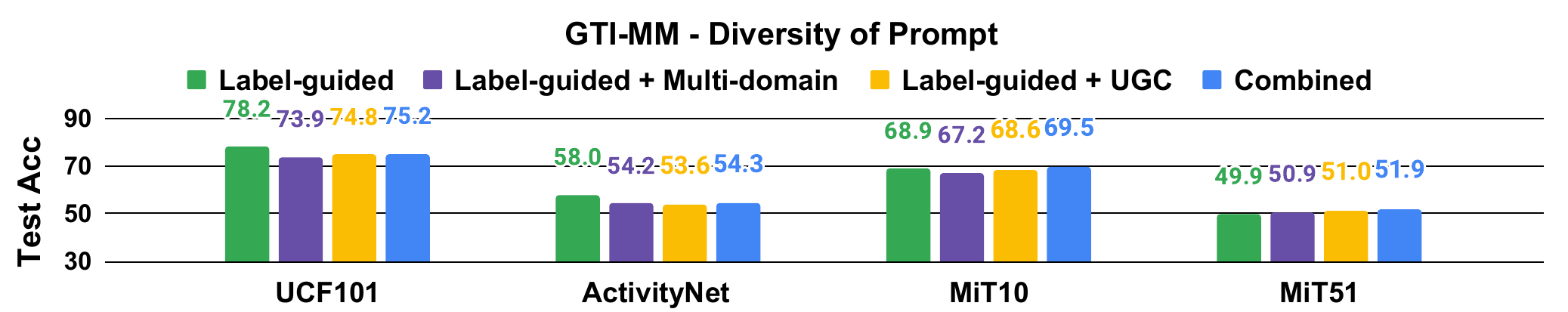}
    \caption{Impact of generation tricks on \texttt{GTI-MM} performance with MiT10 dataset added.}
    \label{fig:tricks_sup}
\end{figure*}

\begin{figure*}[t]
	\centering
	\includegraphics[width=0.9\linewidth]{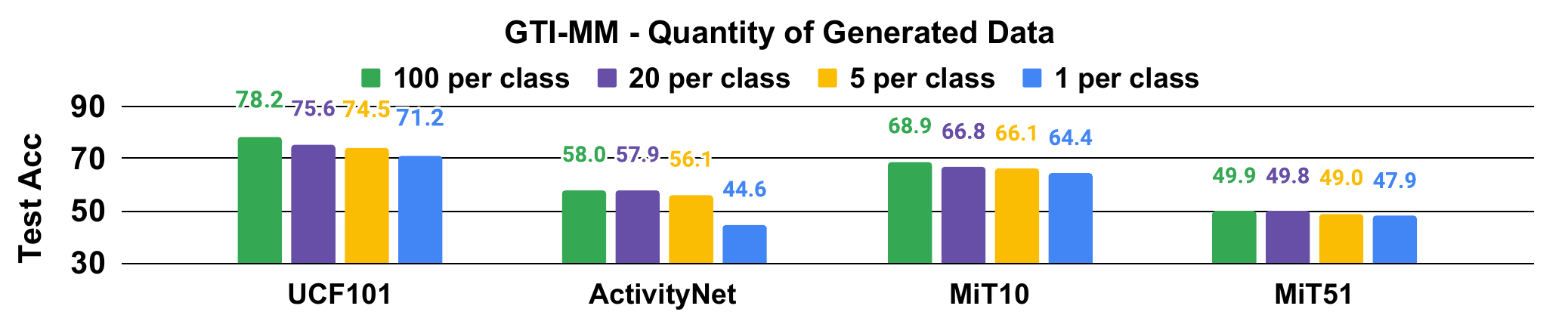}
    \caption{Impact of generation quantity on \texttt{GTI-MM} performance with MiT10 dataset added.}
    \label{fig:quantity_sup}
\end{figure*}

\begin{figure}[t]
	\centering
	\includegraphics[width=0.6\linewidth]{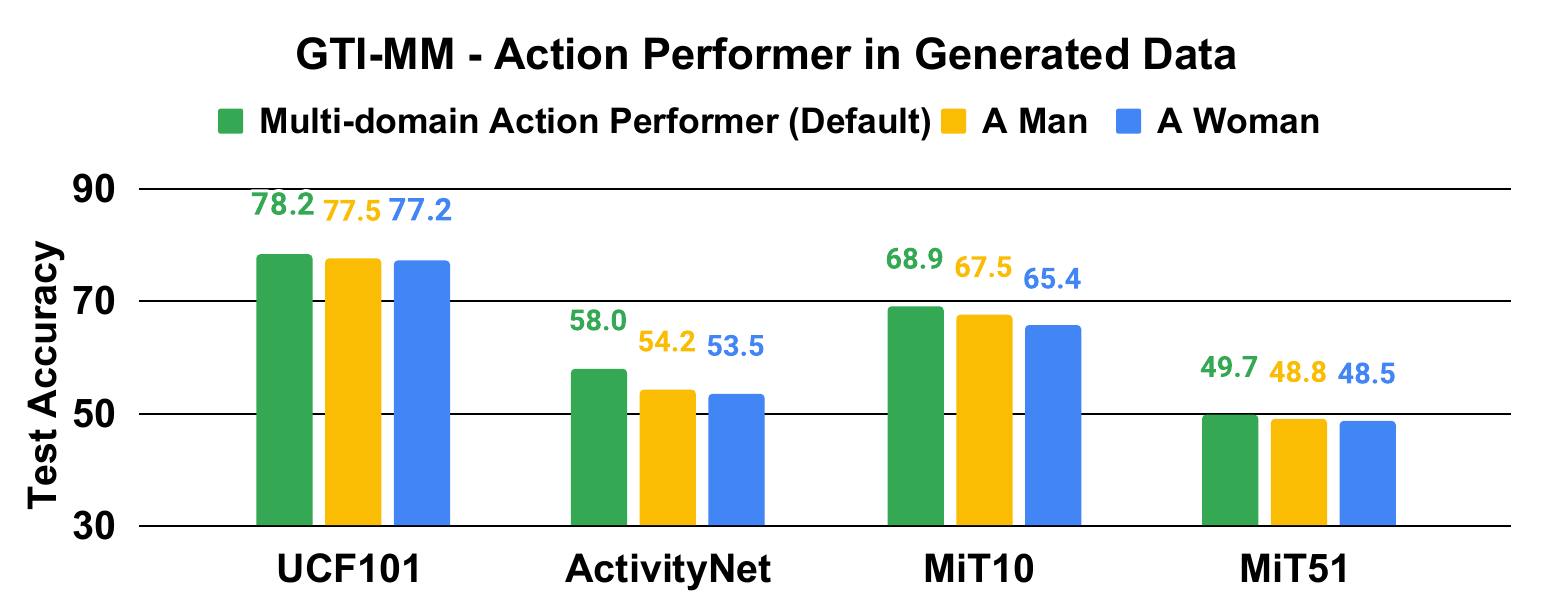}
    \caption{Impact of varying action performers in image generation on \texttt{GTI-MM} performance with MiT10 dataset added.}
    \label{fig:action_performer_sup}
\end{figure}

\section{Supplement Experimental Details}

\subsection{Experimental Details on Text-Visual Datasets}
Similar to audio-visual experiments, We perform the experiments on text-visual datasets three times with different seeds and report the average performance. We report the test performance based on the best performance on the validation set during the training. We evaluate Crisis-MMD and UPMC Food101 using F1 and top-1 accuracy, respectively. We adopt the CLIP model with the ViT visual encoder \cite{radford2021learning} as the backbone for text-visual experiments.

\subsection{Hyperparameter Settings}
To determine the optimal hyperparameters, we conducted a search within specified ranges. The learning rate was searched in \{0.0001, 0.0002, 0.0005\}. We apply a weight decay in the range of 1e-4, input batch size of 128, and epoch number of 15 and 30 for audio-visual and text-visual datasets, respectively. The specific hyperparameter selections for the audio-visual and text-visual experiments are provided below in Table~\ref{hyper:av_experiments} and Table~\ref{hyper:tv_experiments}, respectively.

\section{Supplement Results With MiT10 Datasets}

\subsection{Diversity of Visual Generation}

We provide additional results with MiT10 datasets on the performance of GTI-MM varying action performers in Figure~\ref{fig:action_performer_sup}. The results provide additional evidence that incorporating multi-domain information to action performers enhances the model performance. Moreover, we supply complete performance comparisons regarding the generation tricks as demonstrated in Figure~\ref{fig:tricks_sup}. The results on MiT10 datasets support that combining generation tricks in data generation yields better performances in \texttt{GTI-MM} when data in downstream tasks involve multi-domain knowledge.

\begin{table*}[t]
    \caption{Performance comparisons between \texttt{GTI-MM} and the baseline in text-visual datasets with low-resource visual data presented in training. The testing data are with complete modalities.}
    \vspace{-1mm}
    \centering
    \small

    \resizebox{0.9\linewidth}{!}{
    \begin{tabular}{lcccccc}
        \toprule
        & 
        \multicolumn{1}{c}{\textbf{Training Visual}} &
        \multirow{2}{*}{\textbf{Metric}} &
        \multicolumn{1}{c}{\textbf{Zero-filling Imputation}} & 
        \textbf{\texttt{GTI-MM}} \\
        
        & 
        \multicolumn{1}{c}{\textbf{Missing Ratio ($p$)}} & &
        \multicolumn{1}{c}{\textbf{Multimodal Training}} & \textbf{(Ours)}\\

        \midrule 
        \textbf{Crisis-MMD} & $95\%$ & F1 & $29.51\pm1.61$ & $\mathbf{41.52\pm0.49}$ \\
        \textbf{UMPC Food101} & $95\%$ & Top3-Acc & $87.48\pm0.13$ & $\mathbf{90.36\pm0.28}$ \\

        \bottomrule
    \end{tabular}
    }
\vspace{-2mm}
\label{tab:gti_mm_tv_baseline}
\end{table*}

\begin{figure}[t] {
    \centering
    
    \begin{tikzpicture}

        \node[draw=none,fill=none] at (0,0){\includegraphics[width=0.5\linewidth]{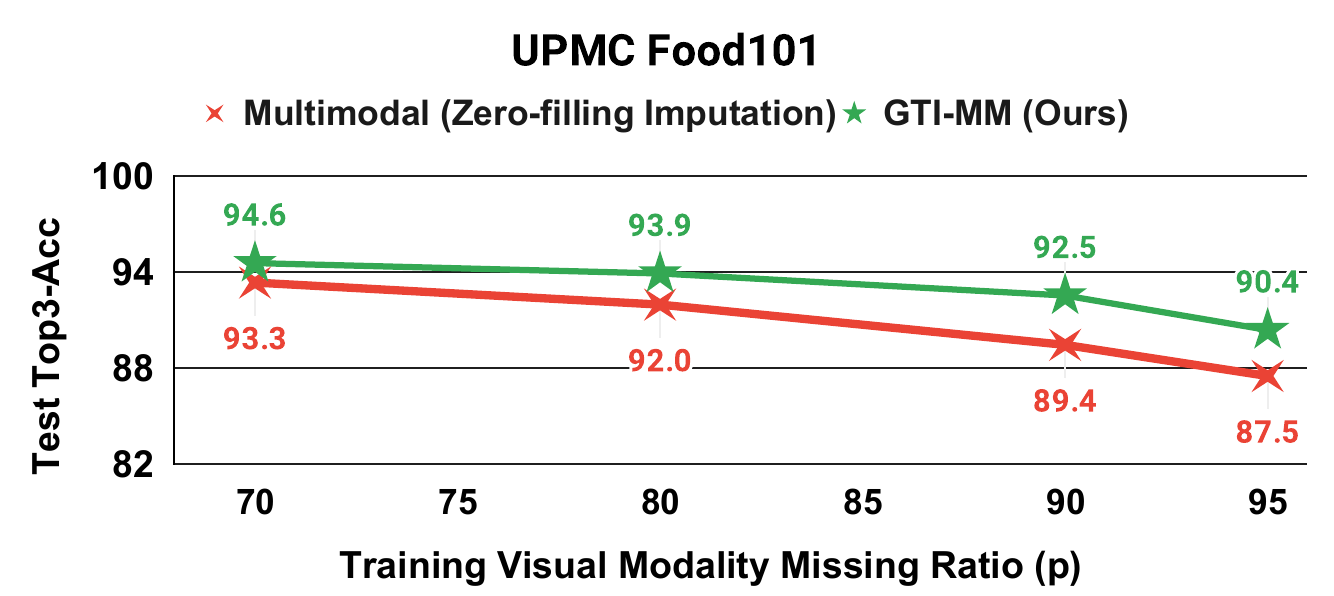}};

        \node[draw=none,fill=none] at (0.5\linewidth,0){\includegraphics[width=0.5\linewidth]{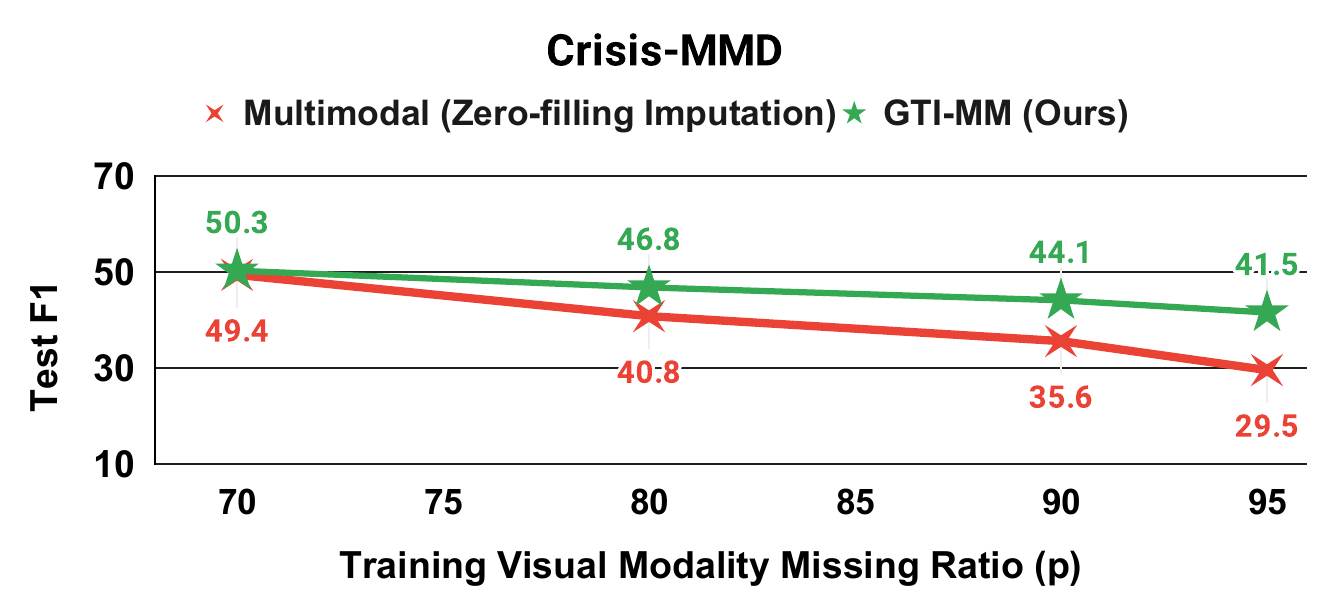}};

    \end{tikzpicture}
    \vspace{-3mm}
    \caption{Comparisons among \texttt{GTI-MM} and baseline at different training visual modality ratios $p$ on text-visual datasets. The testing data are with complete visual and text modalities.}
    \label{fig:missing_ratio_tv}
} \end{figure}

\begin{table}[t]
    \caption{Comparisons between MM-Dropout (Multimodal-Dropout) and \texttt{GTI-MM Dropout}. $p$ and $q$ are training and testing visual missing ratios, respectively.}
    \centering
    \footnotesize
    \begin{tabular*}{0.68\linewidth}{cccccc}
        \toprule
        \multirow{2}{*}{\textbf{Dataset}} & 
        \multirow{2}{*}{\textbf{Method}} &
        \multirow{2}{*}{$\mathbf{p}$} &
        \multicolumn{3}{c}{\textbf{Test Missing Ratio ($q$)}} \\
        
        & & &
        \multicolumn{1}{c}{\textbf{50}} &
        \multicolumn{1}{c}{\textbf{70}} &
        \multicolumn{1}{c}{\textbf{90}} \\

        \cmidrule(lr){1-6}
        \multirow{3}{*}{\textbf{Crisis-MMD}} & \textbf{MM-Dropout} & $0$ & $38.5$ & $32.5$ & $27.6$ \\

        \cline{2-3} & \multirow{3}{*}{\textbf{GTI-MM Dropout}} & $50$ & $\mathbf{38.8}$ & $\mathbf{33.1}$ & $\mathbf{29.1}$  \rule{0pt}{2.25ex} \\
        & & $90$ & $34.2$ & $31.3$ & $29.0$ \\
        & & $95$ & $33.2$ & $30.7$ & $28.7$ \\

        \cmidrule(lr){1-6}
        
        \multirow{3}{*}{\textbf{MM-Food101}} & \textbf{MM-Dropout} & $0$ & $91.0$ & $89.1$ & $87.5$ \\

        \cline{2-3}
        & \multirow{3}{*}{\textbf{GTI-MM Dropout}} & $50$ & $\mathbf{91.1}$ & $\mathbf{89.4}$ & $\mathbf{88.0}$ \rule{0pt}{2.25ex} \\
        & & $90$ & $90.6$ & $89.1$ & $87.9$ \\
        & & $95$ & $90.2$ & $88.9$ & $87.7$ \\

        \bottomrule
    \end{tabular*}
    \label{tab:dropout_gti_mm_tv}
\end{table}

\subsection{Quantity of Visual Generation}

We report additional results with MiT10 datasets on the performance of GTI-MM varying action performers in Figure~\ref{fig:quantity_sup}. The results show that the performance of \texttt{GTI-MM} is robust, requiring only 5 image generations per class to achieve competitive performance compared to 100 image generations per class.

\section{Results With Text-visual Datasets}

\subsection{Can \textbf{\texttt{GTI-MM}} improve data efficiency with training visual modality missing?}

\noindent \textbf{Baseline Comparisons} Table~\ref{tab:gti_mm_tv_baseline} presents the performance comparisons between \texttt{GTI-MM} and the baseline of the multimodal learning with zero-filling imputation on visual modality. We choose this baseline as it exhibits the best performance in audio-visual experiments. Similar to audio-visual experiments, \texttt{GTI-MM} employs the label prompt to generate visual data, producing 100 images per class category. Likewise, the results demonstrate that our proposed multi-modal learning approach \texttt{GTI-MM} can extend to text-visual datasets in visual recognition with visual modality missing in training data. Our results show that \texttt{GTI-MM} substantially improves the multi-modal models from the baseline approach across datasets by imputing missing visual data with synthetic images. 

\vspace{1mm}
\noindent \textbf{Is \texttt{GTI-MM} effective when $p$ is low?} We conduct similar experiments on text-visual datasets to study the effectiveness of \texttt{GTI-MM} at different $p$, as reported in Section~\ref{subsec:comparing_p}. Specifically, we plot the performance of \texttt{GTI-MM} at different visual-modality missing ratios in training, as demonstrated in Figure~\ref{fig:missing_ratio_tv}. The plot exhibits that as the visual modality missing ratio decreases, the performance differences between the low-resource visual model and \texttt{GTI-MM} are reasonably close, and our proposed \texttt{GTI-MM} can still yield better performances than the baseline approach even when $p=70\%$.

\subsection{Can \textbf{\texttt{GTI-MM}} improve model robustness with visual modality missing?}

We further investigate the efficacy of \texttt{GIT-MM} in mitigating visual modality missing in both training and testing data within text-visual applications. We follow our experimental procedures in Section~\label{sec:mm_any_data}. The results of visual modality missing in text-visual datasets are shown in Table~\ref{tab:dropout_gti_mm_tv}. Through comparisons, we observe that \texttt{GTI-MM}, even with $p=95\%$, exhibits competitive performance when encountering visual modality missing in testing data compared to complete multi-modal dropout training. Moreover. as the testing visual missing ratio $q$ increases, \texttt{GTI-MM} demonstrates larger performance improvements compared to multi-modal dropout training with complete data. This suggests that incorporating random sampling of generated content enhances dropout training by introducing more randomness into the training process. Furthermore, the results show that when the training missing ratio reaches $p=50\%$, \texttt{GTI-MM} consistently outperforms multi-modal dropout training with complete data at different levels of $q$. This further demonstrates the effectiveness of \texttt{GTI-MM} in improving modeling robustness, even in scenarios where visual modality is missing in both training and testing data.

\begin{figure*}[t]
	\centering
	\includegraphics[width=0.7\linewidth]{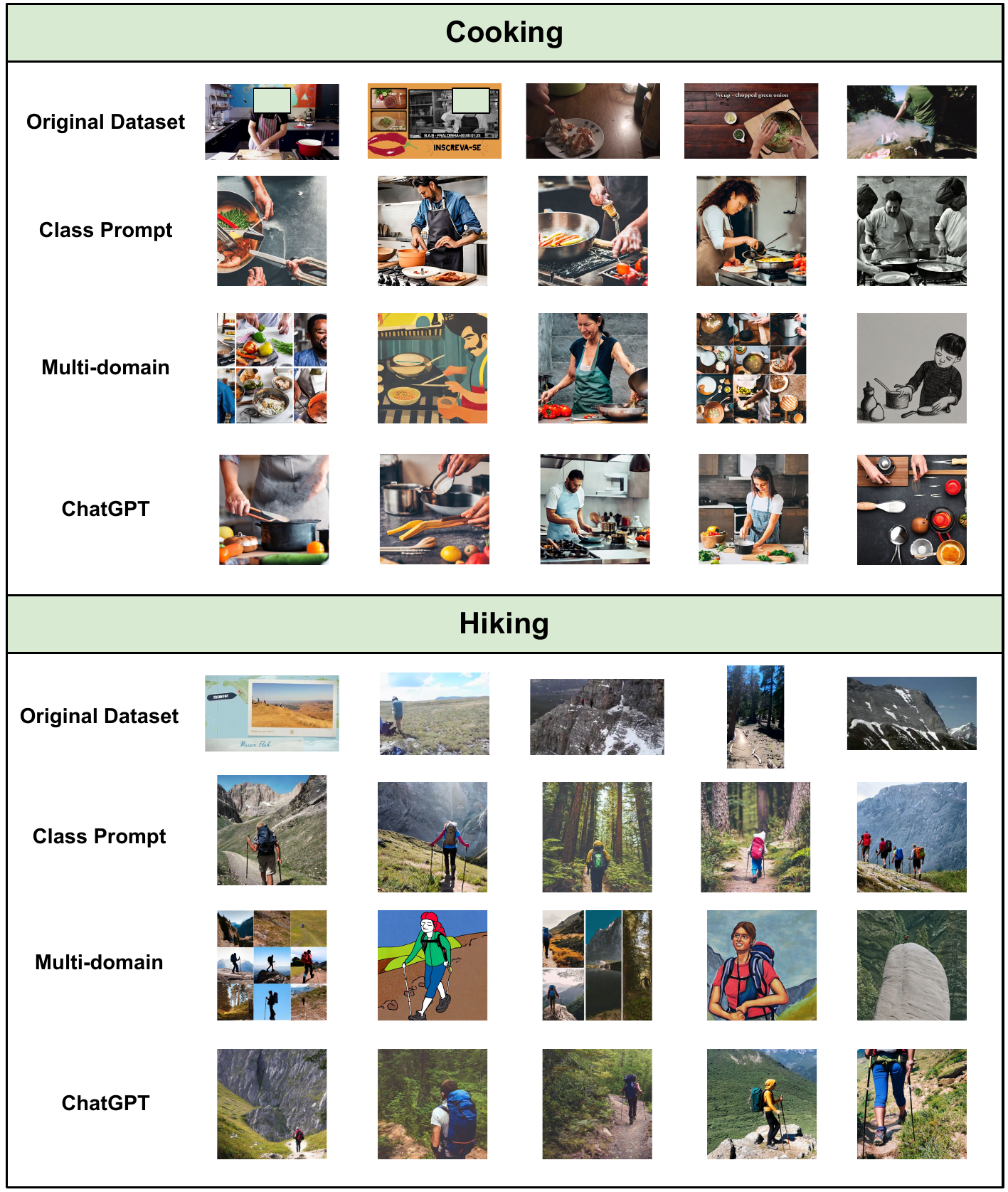}
    \caption{Visual generation examples for classes of cooking and hiking.}
    \label{fig:demo}
\end{figure*}

\section{Visual Generation Examples}

Here, we provide the visual generation examples using the class, multi-domain, ChatGPT-assisted prompt, and original datasets, as demonstrated in Figure~\ref{fig:demo}. We show examples of the original data from the MiT dataset.




